\documentclass[11pt,a4paper]{article}
\usepackage[hyperref]{naaclhlt2019}
\usepackage{times}
\usepackage{latexsym}
\usepackage{tabularx}
\usepackage{multirow}
\usepackage{url}
\usepackage{enumitem}
\usepackage{graphicx}
\usepackage{nccmath}
\usepackage{booktabs}% http://ctan.org/pkg/booktabs
\newcommand{\tabitem}{~~\llap{\textbullet}~~}
\usepackage{url}
\usepackage{float}
\aclfinalcopy % Uncomment this line for the final submission
\def\aclpaperid{279} %  Enter the acl Paper ID here

\title{Conversation Model Fine-Tuning for \\ Classifying Client Utterances in Counseling Dialogues}

\author{Sungjoon Park \textsuperscript{1},~~Donghyun Kim \textsuperscript{2},~~Alice Oh \textsuperscript{1}\\
  \textsuperscript{1} School of Computing, KAIST, Republic of Korea \\
  \textsuperscript{2} Trost, Humart Company, Inc., Republic of Korea \\
  {\tt sungjoon.park@kaist.ac.kr, kdhyeon821@gmail.com} \\
  {\tt alice.oh@kaist.edu} 
  \\}

\date{}

\begin{document}
\maketitle
\begin{abstract}
The recent surge of text-based online counseling applications enables us to collect and analyze interactions between counselors and clients. 
A dataset of those interactions can be used to learn to automatically classify the client utterances into categories that help counselors in diagnosing client status and predicting counseling outcome. 
With proper anonymization, we collect counselor-client dialogues, define meaningful categories of client utterances with professional counselors, and develop a novel neural network model for classifying the client utterances. The central idea of our model, ConvMFiT, is a pre-trained conversation model which consists of a general language model built from an out-of-domain corpus and two role-specific language models built from unlabeled in-domain dialogues. The classification result shows that ConvMFiT outperforms state-of-the-art comparison models. Further, the attention weights in the learned model confirm that the model finds expected linguistic patterns for each category.
\end{abstract}

\section{Introduction}

% para 1
%Mental disorders have become a global health issue. The global population with depression is estimated to be 4.4\%, about 322 million people, and people with anxiety disorder numbers 264 million. \cite{world2017depression}. 
Some mental disorders are known to be treated effectively through psychotherapy. However, people in need of psychotherapy may find it challenging to visit traditional counseling services because of time, money, emotional barriers, and social stigma \cite{bearse2013barriers}. Recently, technology-mediated psychotherapy services emerged to alleviate these barriers. Mobile-based psychotherapy programs \cite{mantani2017smartphone}, fully automated chatbots \cite{ly2017fully, fitzpatrick2017delivering}, and intervention through smart devices \cite{torrado2017emotional} are examples. Among them, text-based online counseling services with professional counselors are becoming popular because clients can receive these services without traveling to an office and with reduced financial burden compared to traditional face-to-face counseling sessions~\cite{talkspace}. 
%A positive effect of these services is that  researchers can collect these counselor-client interactions and  analyze them computationally.

In text-based counseling, the communication environment changes from face-to-face counseling sessions. 
The counselor cannot read non-verbal cues from their clients, and the client uses text messages rather than spoken utterances to deliver their thoughts and feelings, resulting in changes of dynamics in the counseling relationship. Previous studies explored computational approaches to analyzing the dynamic patterns of relationship between the counselor and the client by focusing on the language of counselors \cite{imel2015computational, TACL802}, clustering topics of client issues \cite{dinakar2014real}, and looking at therapy outcomes \cite{howes-purver-mccabe:2014:W14-32, talkspace}.

Unlike previous studies, we take a computational approach to analyze client responses from the \textit{counselor's perspective}. Client responses in counseling are crucial factors for judging the counseling outcome and for understanding the status of the client. So we build a novel categorization scheme of client utterances, and we base our categorization scheme on the cognitive behavioral theory (CBT), a widely used theory in psychotherapy. Also, in developing the categories, we consider whether they are adequate for the unique text-only communication environment, and appropriate for the annotation of the dialogues as training data. Then using the corpus of text-based counseling sessions annotated according to the categorization scheme, we build a novel conversation model to classify the client utterances.

This paper presents the following contributions:

\begin{itemize}
%[noitemsep,topsep=0pt]
\item First, we build a novel categorization method as a labeling scheme for client utterances in text-based counseling dialogues. 
\item Second, we propose a new model, Conversation Model Fine-Tuning (ConvMFiT) to classify the utterances. %It is an extended version of ULMFiT \cite{howard2018universal}, fine-tuning a conversation model grounded on pre-trained language models. 
We explicitly integrate pre-trained language-specific word embeddings, language models and a conversation model to take advantage of the pre-trained knowledge in our model. 
\item Third, we empirically evaluate our model in comparison with other models including a state-of-the-art neural network text classification model. Also, we show typical phrases of counselors and clients for each category by investigating the attention layers. 
\end{itemize}

\begin{table*}[!ht]
\footnotesize
\centering
\bgroup
\def\arraystretch{1.15}%
\begin{tabular}{c|l|l|l|l|l}
\toprule
\begin{tabular}[c]{@{}c@{}} Characteristic~ \end{tabular} & \multicolumn{2}{c|}{\textbf{Informative}} &
\multicolumn{2}{c|}{\textbf{Client Factors}} & \multicolumn{1}{c}{\textbf{Process}} \\ \midrule
\begin{tabular}[c]{@{}c@{}}Category \\ Name\end{tabular} & \multicolumn{1}{c|}{\textbf{\begin{tabular}[c]{@{}c@{}}Factual \\ Information \\ (Fact.)~\end{tabular}}} & \multicolumn{1}{c|}{\textbf{\begin{tabular}[c]{@{}c@{}}Anecdotal \\ Experience \\ (Anec.)\end{tabular}}} & \multicolumn{1}{c|}{\textbf{\begin{tabular}[c]{@{}c@{}}Appealing \\ Problem \\ (Prob.)\end{tabular}}} & \multicolumn{1}{c|}{\textbf{\begin{tabular}[c]{@{}c@{}}Psychological \\ Change \\ (Chan.)\end{tabular}}} & \multicolumn{1}{c}{\textbf{\begin{tabular}[c]{@{}c@{}}Counseling \\ Process \\ (Proc.)\end{tabular}}} \\ \midrule
\begin{tabular}[c]{@{}c@{}}Explanation\end{tabular} & \multicolumn{1}{c|}{\begin{tabular}[c]{@{}c@{}}Brief mention of \\ categorical \\ information\end{tabular}} & \multicolumn{1}{c|}{\begin{tabular}[c]{@{}c@{}}Client’s experience \\ contributing to the \\ appealing problem\end{tabular}} & \multicolumn{1}{c|}{\begin{tabular}[c]{@{}c@{}}Client’s factors \\ related to the \\ appealing problem\end{tabular}} & \multicolumn{1}{c|}{\begin{tabular}[c]{@{}c@{}}~Statement at the \\ resolution stage \\ of the appealing \\ problem~\end{tabular}} & \multicolumn{1}{c}{\begin{tabular}[c]{@{}c@{}}Statement of \\ counseling \\ structure and \\ relationship\end{tabular}} \\ \midrule
\multirow{4}{*}{\begin{tabular}[c]{@{}c@{}}\\[1em]~Examples\end{tabular}} & \begin{tabular}[c]{@{}l@{}}\tabitem Objective \\ ~~~~Fact\end{tabular} & \begin{tabular}[c]{@{}l@{}}\tabitem Experience \\ ~~~~with others\end{tabular} & \begin{tabular}[c]{@{}l@{}}\tabitem Negative \\ ~~~~Emotion\end{tabular} & \begin{tabular}[c]{@{}l@{}}\tabitem Positive \\ ~~~~Prediction\end{tabular} & \begin{tabular}[c]{@{}l@{}}\tabitem A message to \\ ~~~~counselor\end{tabular} \\ 
 & \begin{tabular}[c]{@{}l@{}}\tabitem Living \\ ~~~~conditions\end{tabular} & \begin{tabular}[c]{@{}l@{}}\tabitem Comments \\ ~~~~from others\end{tabular} & \begin{tabular}[c]{@{}l@{}}\tabitem Cognitive \\ ~~~~distortion\end{tabular} & \begin{tabular}[c]{@{}l@{}}\tabitem Expectation,\\ ~~~~Determination\end{tabular} & \begin{tabular}[c]{@{}l@{}}\tabitem Gratitude,\\ ~~~~Greetings\end{tabular} \\
 & \begin{tabular}[c]{@{}l@{}}\tabitem Demographic \\ ~~~~information\end{tabular} & \tabitem Trauma & \begin{tabular}[c]{@{}l@{}}\tabitem Interpersonal \\ ~~~~problems\end{tabular} & \begin{tabular}[c]{@{}l@{}}\tabitem Coping \\ ~~~~behaviors\end{tabular} & \begin{tabular}[c]{@{}l@{}}\tabitem Time \\ ~~~~appointment\end{tabular} \\ 
 & \begin{tabular}[c]{@{}l@{}}\tabitem Limited \\ ~~~~conditions\end{tabular} & \begin{tabular}[c]{@{}l@{}}\tabitem Interpersonal \\ ~~~~situations\end{tabular} & \begin{tabular}[c]{@{}l@{}}\tabitem Family \\~~~~ problems\end{tabular} & \begin{tabular}[c]{@{}l@{}}\tabitem Self-\\ ~~~~awareness\end{tabular} & \begin{tabular}[c]{@{}l@{}}\tabitem Questions about \\ ~~~~the consultation\end{tabular} \\ \bottomrule
\end{tabular}
\egroup
\caption{Final Categorization of client utterances. Five categories are discovered, two for informative giving information to a counselor (Factual information, Anecdotal Experience), two for client factors (appealing problems, psychological change), and the last one for counseling process.}
\label{cats}
\end{table*}

\section{Categorization of Client Utterances}

%\subsection{Objective}
%%%%%%%%%%% categorization is important %%%%%%%%%%%%%%%%
Client responses provide essential clues to understanding the client's internal status which can vary throughout counseling sessions. For example, client's responses describing their problems prevail at the early stage of counseling \cite{hill1978development, replytohill}, but as counseling progresses, problem descriptions decrease while insights and discussions of plans continue to increase \cite{seeman1949study}. Client responses can also help in predicting counseling outcomes. For example, a higher proportion of insights and plans in client utterances indicates a high positive effect of counseling \cite{hill1978development, replytohill}. 
%Thus we try to categorize the client utterances to understand them and to build a labeling scheme for them.

%\subsection{Categorization Objective}
\noindent \textbf{Categorization Objective.} Our final aim is to build a machine learning model to classify the client utterances. Thus, the categorization of the utterances should satisfy the following criteria:

\begin{itemize}
%[noitemsep,topsep=0pt]
\item \textit{Suitable for the text-only environment}: Categories should be detected only using the text response of clients.
\item \textit{Available as a labeling scheme}: The number of categories should be small enough for manual annotation by counseling experts.
\item \textit{Meaningful to counselors}: Categories should be meaningful for outcome prediction or counseling progress tracking.
\end{itemize}

%%%%%%%%%%% why should we construct new categories %%%%%%%%%%%%%%%%
%% Suitable for the text-only environment
\noindent Previous studies in psychology proposed nine and fourteen categories for client and counselor verbal responses, respectively, by analyzing transcriptions from traditional face-to-face counseling sessions \cite{hill1978development, hill1981manual}. But these categories were developed for face-to-face spoken interactions, and we found that for online text-only counseling dialogues, these categories are not directly applicable. Using text without non-verbal cues, a client's responses are inherently different from the transcriptions of verbally spoken responses which include categories such as `silence' (no response for more than 5 seconds) and `nonverbal referent' (physically pointing at a person).
%% Available as a labeling scheme
Another relevant study, derived from text-based counseling sessions with suicidal adolescents proposes 19 categories which we judged to be too many to be practical for manual annotation \cite{responsecounselor2010}.

The last criterion of ``meaningful to counselors" is perhaps the most important. To meet that criterion, we base the categorization process on the \textit{cognitive behavioral theory (CBT)} which is the underlying theory behind psychotherapy counseling. The details of using CBT for the categorization process is explained next. 

%\subsection{Categorization Process}
\noindent \textbf{Categorization Process and Results.}
%%%%%%%%%%% Categorization process: method %%%%%%%%%%%%%%%%
In developing the categories, we follow the Consensual Qualitative Research method \cite{hill1997guide}. Two professional counselors with clinical experience participated in this qualitative research method to define the categorization.
% categorize client utterances within a text-based online counseling session at a general level by referring to other qualitative research methods and case conceptualization studies. Specifically, simplified 

%%%%%%%%%%% Categorization process: stage 1 %%%%%%%%%%%%%%%%
To begin, we randomly sample ten client cases considering demographic information including age, gender, education, job, and previous counseling experiences. 
We then start the categorization process with the fundamental components of the CBT which are \textit{events}, \textit{thoughts}, \textit{emotions}, and \textit{behavior} \cite{hill1981manual}. 
The professional counselors annotate every client utterance to with those initial component categories with tags that add detail. For example, if an utterance is annotated as `emotion', we add     `positive/negative' or concrete label such as `hope'. If these tags are categorized to be a new category, we add that category to the list until it is saturated. When the number of categories becomes more than 40, we define higher level categories that cover the existing categories. 
% As a result, they are finalized in the second stage through discussion.

%%%%%%%%%%% Categorization process: stage 2 %%%%%%%%%%%%%%%%
In the second stage, annotators discuss and merge these categories into five high-level categories. Category 1 is \textit{informative responses to counselors}, and category 2 is \textit{providing factual information and experiences}. Categories 3 and 4 are related to the client factors, \textit{expressing appealing problems} and \textit{psychological changes}. The last category is about the \textit{logistics of the counseling sessions} including scheduling the next session. The categories in detail are as follows:

\begin{itemize}
\item \textit{Factual information (Fact.)} Informative responses to counselor's utterances, including age, gender, occupation, education, family, previous counseling experience, etc.

\item \textit{Anecdotal Experience (Anec.)} Responses describing past incidents and current situations related to the formation of appealing problems. Responses include traumatic experiences, interactions with other people, comments from other people, and other anecdotal experiences.

\item \textit{Appealing Problems (Prob.)} Utterances addressing the main appealing problem which is yet to be resolved, including client's internal factors or their behaviors related to the problems. Specifically, the utterances include cognition, emotion, physiological reaction, and diagnostic features of the problem and desire to be changed.

\item \textit{Psychological Change (Chan.)} Utterances describing insights, cognition of small and big changes in internal factors or behaviors. That is, an utterance at the point where the appealing problem is being resolved.

\item \textit{Counseling Processes (Proc.)} Utterances that include the objective of counseling, requests to the counselor, plans about the counseling sessions, and counseling relationship. This category also covers greetings and making an appointment for the next session.
\end{itemize}

We summarize the category explanations and examples in Table \ref{cats}.
\section{Dataset}
In this section, we explain how counseling dialogues differ from general dialogues, describe the dialogues we collected and annotated, and explain how we preprocessed the data.

\subsection{Characteristics of Counseling Dialogues}
%In this section, we briefly describe unique characteristics of text-based online counseling conversations. 
The counseling dialogues consist of multiple turns taken by a counselor and a client, and each turn can contain multiple utterances. Here we describe two unique characteristics of text-based online counseling conversations compared to general non-goal oriented conversations.

%\subsection{Dialogue Characteristics}
\noindent \textbf{Distinctive roles of speakers.}
Counseling conversations are goal-oriented with the aim to produce positive counseling outcomes, and the two speakers have distinctive roles. The client gives objective information about themselves and subjective experiences and feelings to the counselor to appeal the problems they are suffering from. Then the counselor establishes a therapeutic relationship with the client and elicits various strategies to induce psychological changes in the client. These distinct roles of the conversational participants distinguish counseling conversations from non-goal oriented open-domain response generation modeling.
%which is the language models should be separated by parties.

\begin{figure}[t]
    \centering
    \includegraphics[width=0.45\textwidth]{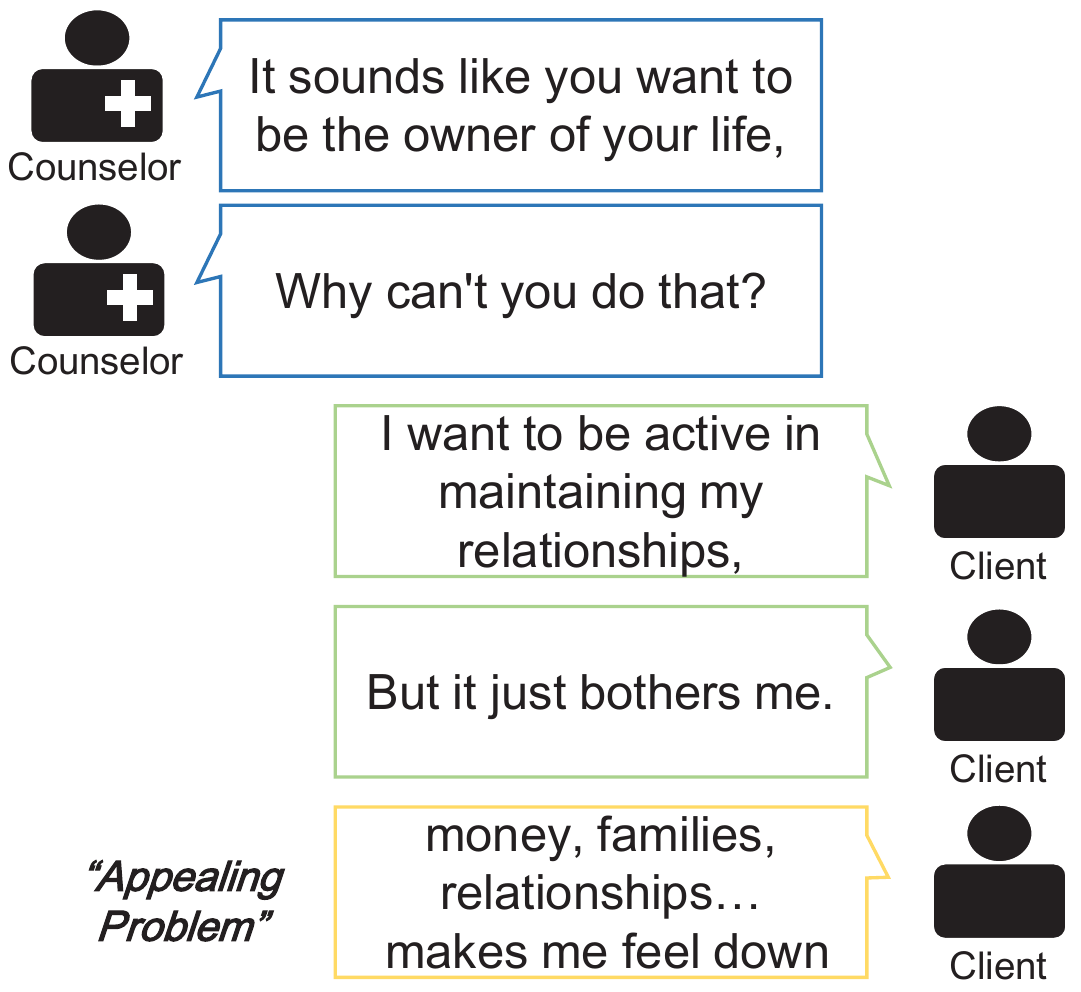}
    \caption{Translated example conversation between a counselor and a client. Each turn can have multiple utterances, and we annotate the client turn at the \textit{utterance}-level. Blue box indicates 1) \textbf{Counselor's utterance}, green is 2) \textbf{Client's context utterance}, and yellow is 3) \textbf{Client's target utterance}. The annotation for the last client utterance is "appealing problem".}
    \label{fig:conversation}
\end{figure}

\noindent \textbf{Multiple utterances in a turn.}
We define an \textit{utterance} as a single text bubble. Counselors and clients can generate multiple utterances in a turn. Especially in a client turn, various information not to be missed by a counselor may occur across multiple utterances. Thus, we treat every utterance separately, as shown in Fig. \ref{fig:conversation}

\subsection{Collected Dataset}
\textbf{Total dialogues.} We collect counseling dialogues of clients with their corresponding professional counselors from Korean text-based online counseling platform Trost. \footnote{https://www.trost.co.kr/} 
Overall, we use 1,448 counseling dialogues which are anonymized before researchers obtain access to the data, by removing personally identifiable information in the dialogues. All meta-data of the dialogues are not provided to the researchers, and named entities such as client's and counselor's names are replaced with random numeric identifiers. The research process including data anonymization and pre-processing is validated by the KAIST Institutional Review Board (IRB).

\noindent \textbf{Labeled Dialogues.} We randomly choose and label 100 dialogues with the discovered five categories. Note that we only label client utterances. Based on these categories, five professional counselors annotated their own client's every utterance in the conversations, as shown in Fig. \ref{fig:conversation}. Note that each utterance can have multiple labels if it includes information across multiple categories.

Table. \ref{conv.stats} shows the descriptive statistics of our labeled dataset. The first two rows present the average lengths of each utterance of counselors and clients in terms of words and characters, showing there is a small difference between the counselor utterance and the client utterance. On the other hand, the average number of utterances in a single counseling session differs; on average, clients write more utterances than counselors.

\begin{table}[h]
\centering
\begin{tabular}{l|r|r|r|r}
\toprule
 & \multicolumn{2}{c|}{Counselor} & \multicolumn{2}{c}{Client} \\ \cmidrule{2-5} 
 & \multicolumn{1}{c|}{\begin{tabular}[c]{@{}c@{}}Mean\end{tabular}} & \multicolumn{1}{c|}{Std.} & \multicolumn{1}{c|}{\begin{tabular}[c]{@{}c@{}}Mean\end{tabular}} & \multicolumn{1}{c}{Std.} \\ \midrule
\begin{tabular}[c]{@{}l@{}}\small{\# of words}\end{tabular} & 6.26 & 6.25 & 5.91 & 14.62 \\ \midrule
\begin{tabular}[c]{@{}l@{}}\small{\# of chars}\end{tabular} & 25.71 & 23.44 & 24.31 & 61.10 \\ \midrule
\begin{tabular}[c]{@{}l@{}}\small{\# of utters}\end{tabular} & 163.2 & 236.97 & 238.72 & 578.49 \\ \bottomrule
\end{tabular}
\caption{Descriptive statistics of labeled counseling dialogues. 
%The length of single utterance is about the same between client and counselors, but the number of client's utterance in a conversation is larger than that of counselors.
}
\label{conv.stats}
\vspace{-4mm}
\end{table}

\subsection{Preprocessing}
We intentionally leave in punctuations and emojis since they can help to infer the categories of the client utterances, treating them as separate tokens. Then we construct triples from the labeled dialogues consisting of 1) counselor's utterances (blue in Fig. \ref{fig:conversation}), 2) client's context utterances (green), and 3) client's target utterances to be categorized. (yellow)

We split the dataset into train, validation, and test sets. Table. \ref{stats.data} shows the number of triples in each set, showing factual information (Fact.) and psychological change (Chan.) categories appear less frequently than the others.

\begin{table}[h]
\centering
\begin{tabular}{l|rrr|r}
\toprule
 & \multicolumn{1}{c}{Train} & \multicolumn{1}{c}{Valid} & \multicolumn{1}{c|}{Test} & \multicolumn{1}{c}{\# of labels} \\ \midrule
Fact. & 817 & 140 & 172 & 1129 \\
Anec. & 5317 & 1180 & 1175 & 7726 \\
Prob. & 4728 & 1004 & 981 & 6713 \\
Chan. & 887 & 211 & 199 & 1297 \\
Pros. & 4570 & 964 & 961 & 6459 \\ \midrule
\begin{tabular}[c]{@{}l@{}}\small{\# of triples}\end{tabular} & 14679 & 3166 & 3165 & 21100 \\ \bottomrule
\end{tabular}
\caption{The number of triples (partner's utterance, client's context utterance, client's target utterance) and corresponding labels for each set. 
%In total, train, valid, and test set includes 14,679, 3,166, 3,165 triples, respectively. 
Factual information (Fact.) and psychological change (Chan.) categories have less number of instances compared to the others.}
\label{stats.data}
\vspace{-4mm}
\end{table}

\begin{figure*}[!ht]
    \centering
    \includegraphics[width=\textwidth]{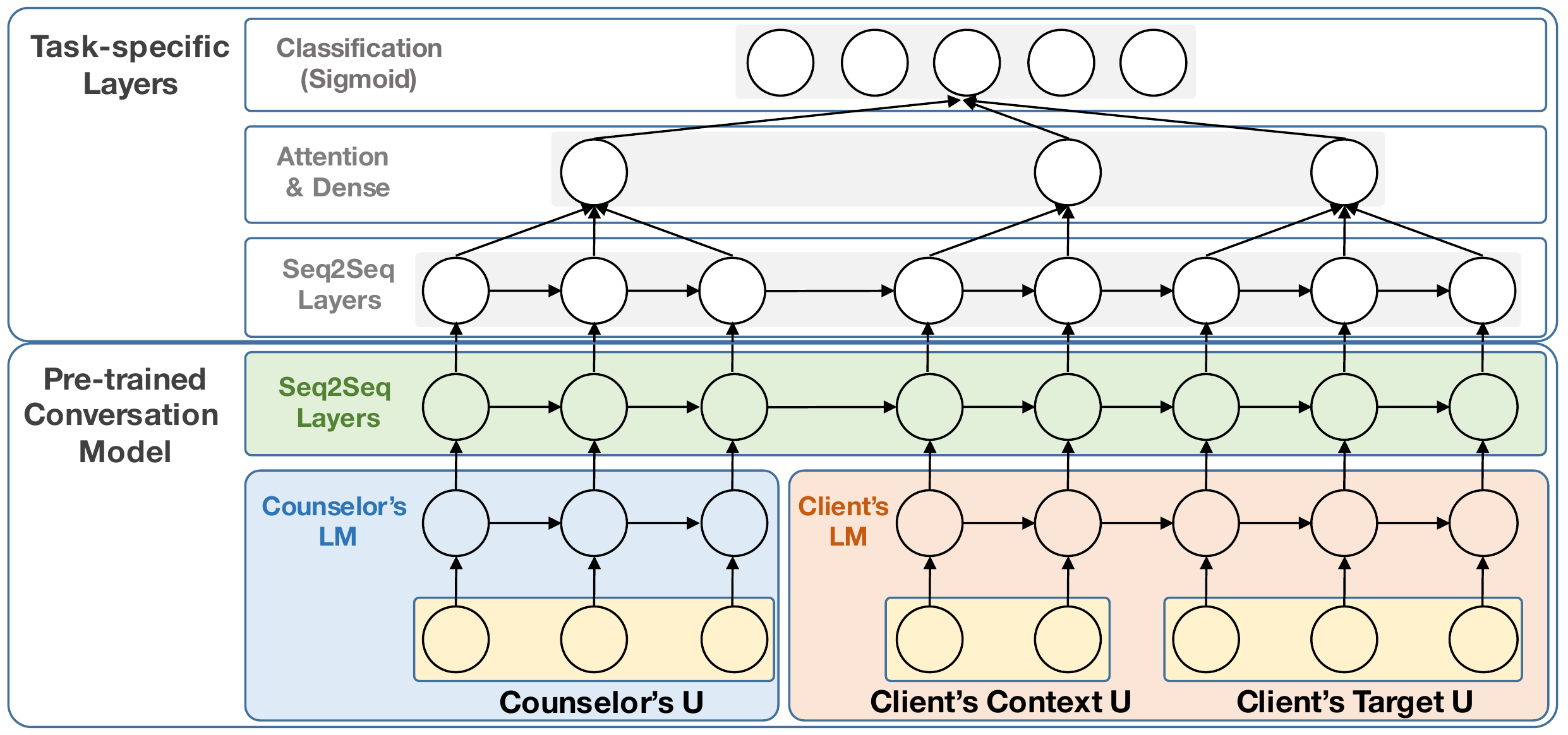}
    \caption{ConvMFiT (Conversation Model Fine-Tuning) model architecture. We use a pre-trained conversation model using seq2seq models. The model is based on pre-trained counselor's and client's language models. (lower colored part) Then, we stack additional task-specific seq2seq layers and classification layers over the conversation model to learn specific features. Between the two layers, attention layer is added to enhance its interpretability.}
    \label{fig:model}
\end{figure*}

\section{Model}
We introduce ConvMFiT (Conversation Model Fine-Tuning), fine-tuning pre-trained seq2seq based conversation model to classify the client's utterances.

\noindent \textbf{Background.} Our corpus of 100 labeled conversations is not large enough to fully capture the linguistic patterns of the categories without external knowledge. The small size of the dataset is difficult to overcome because the labeling by professional counselors is costly. We found a potentially effective solution for a small labeled dataset by using pre-trained language models for various NLP tasks \cite{ramachandran-liu-le:2017:EMNLP2017, howard2018universal}. Therefore, we focus on transferring knowledge from unlabeled in-domain dialogues as well as a general out-of-domain corpus. 

%\subsection{Model Overview}
\noindent \textbf{Overview.} 
We illustrate the overall model architecture. We first stack pre-trained seq2seq layers which represent a conversation model (see Fig. \ref{fig:model}, lower colored part). Then we stack additional seq2seq layers and classification layers over it to capture task-specific features, and an attention layer is added between the two to enhance the model's interpretability (see Fig. \ref{fig:model}, upper white part).

This approach leverages the advantages of using a language model based conversation model for transfer learning. The model can be regarded as an extended version of ULMFiT \cite{howard2018universal} with modifications to fit our task. In ConVMFiT, the model accepts a pre-trained seq2seq conversation model that requires two pre-trained language models, like an encoder and a decoder, learning the  dependencies between them on the seq2seq layers. This approach is shown to be effective for machine translation, applying a source language model for the encoder and target language model for the decoder  \cite{ramachandran-liu-le:2017:EMNLP2017}.
%In our model, counselor's language model can be regarded as encoder, and client's language model as decoder. 
%Like ULMFiT, As shown in Fig. \ref{fig:model}, task-specific layers are added on the top of pre-trained model.

\subsection{Model Components}
\noindent \textbf{Word Vectors.} Counseling dialogues consist of natural Korean text which is morphologically rich, so we train word vectors specifically developed for the Korean language \cite{P18-1226}. We train the vectors over a Korean corpus including out-of-domain general documents, 1) Korean Wikipedia, 2) online news articles, and 3) Sejong Corpus, as well as in-domain unlabeled counseling dialogues. The corpus contains 0.13 billion tokens. To validate the trained vector quality, we check performance on word similarity task (WS353) for Korean \cite{P18-1226}. The Spearman’s correlation is 0.682, which is comparable to the state-of-the-art performance. These vectors are used as inputs (see Fig. \ref{fig:model}, \textcolor{yellow}{yellow}).

\noindent \textbf{Pre-trained Language Models.} We assume that counselors and clients have different language models because they have distinctive roles in the dialogue. 
%Counseling aims to help client resolving their appealing problems, so trained professional counselors have various conversational strategies to help clients and utilize them during conversation. However, clients describe their issues with plain language. 
Therefore, we train a counselor language model and a client language model separately. We collect counselor utterances in the total dialogue dataset, except dialogues in the test set, to train a counselor language model, and all others are used for training a client language model. Then, We fine-tune the two trained LMs with utterances in the labeled dialogues.

For each model, we train word-level language models by using multilayer LSTMs. We apply various regularization techniques, weight tying \cite{Hakan2017tying}, embedding dropout and variational dropout \cite{Gal2016Theoretically}, which are used to regularize LSTM language models \cite{Hakan2017tying, ramachandran-liu-le:2017:EMNLP2017, merity2018regularizing}.

We use a 3-layer LSTM model having 300 hidden units for every layer. We set embedding dropout and output dropout for each layer to 0.2, 0.1, respectively. The pre-trained language models generate inputs to seq2seq layer of a conversation model (Fig. \ref{fig:model}, \textcolor{blue}{blue} and \textcolor{red}{red}).

\noindent \textbf{Pre-trained Conversation Model.} Next, we train a seq2seq conversation model \cite{vinyals2015neural}. We use the pre-trained counselor language model as an encoder, and the client language model as a decoder. The dependency of the decoder on the encoder is trained by seq2seq layers, stacked over the pre-trained models. (Fig. \ref{fig:model}, \textcolor{green}{green})

We stack 2-layer LSTMs over the pre-trained counselor and client language models, respectively. The final states of the LSTMs on the counselor language model is used as an initial state of the LSTMs on the client language model. We set the hidden unit size of 300 for every LSTMs and set output dropout to 0.05. The outputs of pre-trained conversation models is used for inputs to seq2seq layers of the task-specific layers.

During training the conversation model, we regularize the parameters of the model by adding cross entropy losses of the pre-trained counselor and client language models to seq2seq cross entropy loss of the conversation model. Three losses are weighted equally. This prevents catastrophic forgetting of the pre-trained language models and is important to achieve high performance  \cite{ramachandran-liu-le:2017:EMNLP2017}.
%We leave the exploration of more complex architecture of the conversation model integrated with pre-trained language models, expecting it might increase performance if it could model conversations better.
Also, there is room for improvement of the architecture of a conversation model, which integrates pre-trained language models, to capture dialogue patterns better and thus leads to higher classification performance. We will explore the architecture in future work.

\noindent \textbf{Task-specific layers.}
By leveraging pre-trained language model based conversation model, we finally add layers for classification. In order to capture task-specific features, we first stack seq2seq layers over the conversation model. Then we add attention mechanism for document classification \cite{yang-EtAl:2016:N16-13}. Lastly, we use a sigmoid function as an output layer to predict whether the information is included in utterances because multiple categories can appear in a single utterance. (Fig. \ref{fig:model}, \textcolor{gray}{gray})

We use a 2-layer LSTM model for the seq2seq layers. We set 300 hidden units for every LSTMs and set output dropout to 0.05. The size of attention layer is set to 500. 

\subsection{Model Training}
Thus the model is trained by three steps: 1) training word vectors and two language models, 2) training seq2seq conversation model with pre-trained LMs, and 3) fine-tuning task-specific classification layers, after removing softmax of the conversation model. In the last step, we compute binary logistic loss between predicted probability for each category and label as a loss function. We use Adam with default parameters for the optimizer, in order to train language models, conversation model, and fine-tuning the classifier. Also, gradual unfreezing is applied while training the model, starting updating parameters from the task-specific layers and unfreezing the next lower frozen layer for each epoch. We unfreeze layers every other epoch until all layers are tuned, and we stop training when the validation loss is minimized. All hyperparameters are tuned over the development set.

\section{Experiments}
\subsection{Comparison Models}
We compare our model with baseline models in Table \ref{table:quan}. Models 1-5 are classifiers which only use the target client utterance to classify it, and models 6-8 are conversation model-based classifiers considering counselor's utterances and client's context utterances. All models use the same pre-trained word vectors for a fair comparison.

%\vspace{0.2em}
\noindent \textbf{(1) Random Forest, (2) SVM with RBF kernel.} We represent an utterance by an average of all word vectors in it, then feed it to the classifier input.

%\vspace{0.2em}
\noindent \textbf{(3) CNN for text classification} \cite{kim:2014:EMNLP2014}. In the convolution layer, we set the filter size to 1-10, and 30 filters each, then we apply max-over-time pooling. Then, a dense layer and sigmoid activation is applied over the layer.
%\cite{collobert2011natural}

%\vspace{0.2em}
\noindent \textbf{(4) RNN}
Bidirectional LSTM is used where the final states for each direction are concatenated to represent the utterances. Then, a dense layer and a sigmoid activation are stacked.
%, and in (5), we add an attention layer for document classification \cite{yang-EtAl:2016:N16-13}.

%\vspace{0.2em}
\noindent \textbf{(5) ULMFiT.} A pre-trained client language model is used as a universal language model. Details are the same as described in Section 5.3. In addition, 2-LSTM layers and dense layer with sigmoid activations are stacked over the LM for classification. Gradual unfreezing is applied during training \cite{howard2018universal}.

%\vspace{0.2em}
\noindent \textbf{(6) Seq2Seq.} Like encoder-decoder based conversation model \cite{vinyals2015neural}, three LSTMs are assigned for each Counselor's utterances, client's context/target utterances. The initial states of the client language models are set to the final states of preceding utterances. Then dense \& sigmoid layers for classification are stacked over the final state of the client's target utterances.

%\vspace{0.2em}
\noindent \textbf{(7) HRED.} Hierarchical encoder-decoder model (HRED) is used as a conversation model \cite{serban2016building}. For the encoder RNN, counselor's utterances and client's context utterances are given as inputs, and their information is stored in context RNN, which delivers it to the decoder accepting client's target utterances. Like (6) Seq2Seq, dense \& sigmoid layers for classification are stacked over the final state of the client's target utterances.

\subsection{Ablation Study}
We conduct an ablation study to investigate the effect of the pre-trained models in Table \ref{table:quan_trl}.  

%\vspace{0.2em}
\noindent \textbf{(1-4) Adding pre-trained models.} Model 1-4 have the same architecture as ConvMFiT, which is Model (8) in Table. \ref{table:quan}. Model (1) in Table. \ref{table:quan_trl} initializes every parameter randomly. Model (2) starts training only with pre-trained word vectors. Model (3) leverages counselor and client language models as well, and Model (4) shows the performance of ConvMFiT. As Model (3) and (4) use more than two pre-trained components, gradual unfreezing in applied, unfreezing shallower layers first during training.

%\vspace{0.2em}
\noindent \textbf{(4-1) Task-specific Seq2Seq Layers.} Model (4-1) removes task-specific Seq2Seq layers in the model, which leaves only attention and dense layer to capture task-specific features. It may result in an insufficient model capacity to capture relevant features for the task.

%\vspace{0.2em}
\noindent \textbf{(4-2) Effect of Gradual Unfreezing.} Model (4-2) is trained without gradual unfreezing, allowing the parameters of every layer in the model change by the gradients from the first epoch.

\begin{table*}[!ht]
\centering
\begin{tabular}{c|c|l|rrrrr|rr|r}
%\hline
\toprule
\small{No.} & \small{Dep.} & \multicolumn{1}{c|}{Model} & \multicolumn{1}{c}{\begin{tabular}[c]{@{}c@{}}F1\\ (Fact.)\end{tabular}} & \multicolumn{1}{c}{\begin{tabular}[c]{@{}c@{}}F1\\ (Anec.)\end{tabular}} & \multicolumn{1}{c}{\begin{tabular}[c]{@{}c@{}}F1\\ (Prob.)\end{tabular}} & \multicolumn{1}{c}{\begin{tabular}[c]{@{}c@{}}F1\\ (Chan.)\end{tabular}} & \multicolumn{1}{c|}{\begin{tabular}[c]{@{}c@{}}F1\\ (Proc.)\end{tabular}} & \multicolumn{1}{c}{\begin{tabular}[c]{@{}c@{}}Macro \\ Prec\end{tabular}} & \multicolumn{1}{c|}{\begin{tabular}[c]{@{}c@{}}Macro \\ Rec\end{tabular}} & \multicolumn{1}{c}{\begin{tabular}[c]{@{}c@{}}Macro \\ F1\end{tabular}} \\ \midrule
1 & X & RF & .000 & .564 & .420 & .000 & .723 & .476 & .269 & .341 \\
2 & X & SVM(rbf) & .012 & .683 & .457 & .000 & .766 & .602 & .385 & .384 \\
3 & X & CNN & .211 & .528 & .506 & .128 & .706 & .450 & .397 & .416 \\
4 & X & RNN & .193 & .574 & .570 & .046 & .770 & .607 & .375 & .431 \\
5 & X & ULMFiT & .205 & .641 & .591 & .057 & .784 & .613 & .413 & .455 \\ \midrule
6 & O & Seq2Seq & .263 & .662 & .678 & .226 & .823 & .695 & .472 & .530 \\
7 & O & HRED & .261 & .706 & .675 & .193 & .820 & .680 & .475 & .531 \\
8 & O & ConvMFiT & \textbf{.441} & \textbf{.761} & \textbf{.726} & \textbf{.447} & \textbf{.835} & \textbf{.716} & \textbf{.602} & \textbf{.642} \\
%\hline
\bottomrule
\end{tabular}
\caption {Classification results. Among the models 1-6 which only use client's utterances to predict categories, (6) ULMFiT show better performance to the others. Models 7-9 are classifiers based on the conversation model, and ConvMFiT outperforms the others (.642).
%because it also accepts conversational context (client's context and counselor's utterances, and using pre-trained conversation model as well.
}
\label{table:quan}
\end{table*}

\begin{table*}[!ht]
\centering
\small
\begin{tabular}{l|ccccc|rrrrr|rrr}
\toprule
\multicolumn{1}{c|}{No.} & Emb. & LM & \begin{tabular}[c]{@{}c@{}} Conv.\\ seq2seq\end{tabular} & \begin{tabular}[c]{@{}c@{}}Grad.\\ Unf.\end{tabular} & \begin{tabular}[c]{@{}c@{}}Task.\\ seq2seq\end{tabular} & \multicolumn{1}{c}{\begin{tabular}[c]{@{}c@{}}F1\\\tiny{(Fact.)}\end{tabular}} & \multicolumn{1}{c}{\begin{tabular}[c]{@{}c@{}}F1\\\tiny{(Anec.)}\end{tabular}} & \multicolumn{1}{c}{\begin{tabular}[c]{@{}c@{}}F1\\\tiny{(Prob.)}\end{tabular}} & \multicolumn{1}{c}{\begin{tabular}[c]{@{}c@{}}F1\\\tiny{(Chan.)}\end{tabular}} & \multicolumn{1}{c|}{\begin{tabular}[c]{@{}c@{}}F1\\\tiny{(Proc.)}\end{tabular}} & \multicolumn{1}{c}{\begin{tabular}[c]{@{}c@{}}Macro\\Prec\end{tabular}} & \multicolumn{1}{c}{\begin{tabular}[c]{@{}c@{}}Macro\\Rec\end{tabular}} & \multicolumn{1}{c}{\begin{tabular}[c]{@{}c@{}}Macro\\F1\end{tabular}} \\ \midrule
1 & X & X & X & - & O & .032 & .706 & .602 & .222 & .711 & .418 & .575 & .455 \\
2 & O & X & X & - & O & .043 & .758 & .661 & .258 & .748 & .459 & .662 & .494 \\
3 & O & O & X & O & O & .425 & .782 & .727 & .365 & .831 & .587 & .719 & .626 \\
4 & O & O & O & O & O & \textbf{.441} & \textbf{.761} & \textbf{.726} & \textbf{.447} & \textbf{.835} & \textbf{.602} & \textbf{.716} & \textbf{.642} \\ \midrule
4-1 & O & O & O & O & X & .307 & .738 & .666 & .304 & .802 & .513 & .687 & .563 \\
4-2 & O & O & O & X & O & .417 & .768 & .721 & .399 & .824 & .591 & .695 & .626 \\
4 & O & O & O & O & O & \textbf{.441} & \textbf{.761} & \textbf{.726} & \textbf{.447} & \textbf{.835} & \textbf{.602} & \textbf{.716} & \textbf{.642} \\ \bottomrule
\end{tabular}
\caption{Ablation study results. All models use the same architecture of ConvMFiT. Adding pre-trained word vectors (2), counselor and client language models (3), seq2seq layers of conversation model (4) results in a performance improvement. 
%Among them, we find pre-trained language models are a crucial factor to the improvement. 
Also, adding task-specific seq2seq layers to ensure the model's sufficient capacity for capturing relevant features (4-1), and applying gradual unfreezing lead to performance improvement (4-2).}
\label{table:quan_trl}
\end{table*}

\section{Results}

\subsection{Classification Performance}
We show the performance of our model and comparison models in Table. \ref{table:quan}. (1) Random Forests and (2) Support Vector Machines underperform to classify utterances correctly which belong to rarely occurred classes.  Compared to (1) and (2), (3) CNN and (4) RNN show better performance since they look at the sequence of words (.416, .431, respectively). RNN shows slightly better performance than CNN. (6) ULMFiT outperforms the others by using a pre-trained client language models (.455).

The client target utterances have their context, and they also depend on the counselor's preceding utterances, so using the preceding counselor utterance as well as the client context utterances helps to improve the classification performance. When integrating the information using simple (6) seq2seq model, it shows better performance (.530) than (5) ULMFiT. (7) HRED adds a higher-level RNN to seq2seq models, but we find there is little performance gain (.001) which makes the model overfit easily. 
%So we omit comparing further models incorporates context RNNs, such as VHRED \cite{serban2017hierarchical}. 

(8) ConvMFiT employs pre-trained conversation model based on pre-trained LMs and so outperforms all other baseline models (.642). This is because ConvMFiT integrates conversational contexts and the counselor language model, which helps to capture the patterns of client's language better. Also, the improvement is higher for the class (Fact.) and (Chan.) which have small numbers of examples since the ConvMFiT could leverage pre-trained knowledge to classify them.

\subsection{Effect of Pre-trained Components}
With an ablation study applying pre-trained components step by step to our model, we show the sources of performance improvement in Table \ref{table:quan_trl}. Model (1) uses the same architecture but no pre-trained models are applied, showing poor performance due to overfitting (.455). The f1 score is lower than that of the simple seq2seq model. Model (2) only uses pre-trained word vectors, and it helps to increase the performance slightly (.494). Also, Model (3) adds two pre-trained LMs with gradual unfreezing, resulting in a substantial performance increase (.626.), so we find pre-trained LMs are essential to the improvement. Lastly, Model (4) adds the dependency between two language models to fully leverage the pre-trained conversation model. This also helps classify utterances better (.642).

In addition, we find only adding attention and dense layers are not sufficient to learn task-specific features, so providing the model more capacity helps improving the performance. Without task-specific seq2seq layers, model (4-1) shows decreased f1 score (.563). Meanwhile, model (4-2) shows careful training scheme affects to the performance as well.

\subsection{Qualitative Results}
To investigate how linguistic patterns of counselors and clients differ in the various categories, we report qualitative results based on the activation values of the attention layer. To protect the anonymity of the clients, we explore key phrases from utterances rather than publish parts of the conversation in any form.

To this end, we compute the relative importance of n-grams. For any $n$ of n-gram in an utterance of length $N$ where $n<N$, the relative importance $r$ is computed as follows:
\setlength{\abovedisplayskip}{0pt}
\setlength{\belowdisplayskip}{0pt}
\begin{equation}
    (~\prod_{i=1}^{n} a_{i} - (~1/N~)^n~)~/~(~1/N~)^n
\end{equation} 

\noindent where $a_{i}$ is corresponding attention value of a word, $\prod_{i=1}^{n} a_{i}$ is the product of the values for every words in the n-gram, normalized by the expected attention weights $(1/N)^n$. The normalization term considers the length of utterance since a word in short utterances tends to have high attention value because of $\sum_{i=1}^{N} a_{i}=1$. 

We name this measure `relative importance' $r$ meaning that the degree of the n-gram is how much it is attended to compared to the expectation. Based on this, we select examples from 100 top-ranked key phrases for each category and the results are presented in Appendix (Table. \ref{table:qual}). All of the presented key phrases are translated to English from Korean.

\noindent \textbf{Factual Information.} Clients provide demographic information and previous experience of visits to counselors or psychiatrists. In some case, clients talk more about the motivation of counseling. Since this information is explored in an early stage of counseling sessions, we also find counselor greetings with client's names.

\noindent \textbf{Anecdotal Experience.} Clients describe their experiences by using past tense verbs. Usually, utterances include phrases such as `I thought that', `I was totally wrong'. Counselors show simple responses `well..', and reflections.

\noindent \textbf{Appealing Problem.} Like anecdotal experience, clients describe their problems, but with using the present tense of verbs. They are appealing their thinking and emotions. Counselors also show simple responses or reflections. Since some clients immediately start pouring out their problems right after counseling sessions starts, so greetings from counselor appear in key phrases.

\noindent \textbf{Psychological Change.} Clients obviously report their change of feelings, emotions or thoughts. It includes looking back on the past and then determining to change in the future. Counselors give supportive responses and empathetic understanding.

\noindent \textbf{Counseling Process.} Clients and counselors exchange greetings with each other. Also, they discuss making an appointment for the next session. In some cases, counselors respond to client's questions about the logistics of the sessions.

%Other works investigate effectiveness of text-based online therapy based on counseling conversations
%progress and conversational outcomes \cite{howes-purver-mccabe:2014:W14-32, talkspace, TACL802}. They apply computational method to large-scale counseling dialogues. From the dialogues, counselor's conversational style \cite{imel2015computational} can be also clustered by using topic models.

%\noindent \textbf{Conversation Modeling.} To develop non-goal oriented conversation models, sequence-to-sequence model is applied to generate response for a given input \cite{vinyals2015neural}. In addition, to consider previous conversations, context information is added to simple seq2seq model as a higher-level recurrent neural network. \cite{serban2016building} To generate various response, stochastic latent variable used. \cite{serban2017hierarchical} Meanwhile, for goal-oriented conversations such as online shopping \cite{yan2017building}, topic models and convolutional neural networks as combined to give users an appropriate response. Especially for QA systems, memory network is proposed \cite{kumar2016ask}. 

\section{Related Work}

%\textbf{Psycho-linguistic Patterns of People in Needs.} 
Researchers have explored psycho-linguistic patterns of people with mental health problems \cite{gkotsis-EtAl:2016:CLPsych1}, depression \cite{resnik-EtAl:2015:CLPsych2}, Asperger's and autism \cite{ji-EtAl:2014:W14-32} and Alzheimer's disease \cite{orimaye-wong-golden:2014:W14-32}. In addition, these linguistic patterns can be quantified, for example, overall mental health \cite{loveys-EtAl:2017:CLPsych, coppersmith-dredze-harman:2014:W14-32}, and schizophrenia \cite{mitchell-hollingshead-coppersmith:2015:CLPsych}.

To aid people with those mental issues, large portion of studies are dedicated to detecting those issues from natural language. Depression \cite{morales-scherer-levitan:2017:CLPsych, jamil-EtAl:2017:CLPsych, fraser-rudzicz-hirst:2016:CLPsych}, 
%obsessive-compulsive disorder (OCD) \cite{roemmele-mardo-gordon:2017:CLPsych}, 
anxiety \cite{shen-rudzicz:2017:CLPsych}, 
%crisis \cite{kshirsagar-morris-bowman:2017:CLPsych}, 
distress \cite{desmet-jacobs-hoste:2016:CLPsych}, and self-harm risk \cite{yates-cohan-goharian:2017:EMNLP2017} can be effectively detected from narratives or social media postings.

\section{Discussion and Conclusion}
In this paper, we developed five categories of client utterances and built a labeled corpus of counseling dialogue. Then we developed the ConvMFiT for classifying the client utterances into the five categories, leveraging a pre-trained conversation model. Our model outperformed comparison models, and this is because of transferring knowledge from the pre-trained models. We also explored and showed typical linguistic patterns of counselors and clients for each category.

Our ConvMFiT model will be useful in other classification tasks based on dialogues. ConvMFiT is a seq2seq model for counselor-client conversation, however, another approach would be to model with existing non-goal oriented conversation models incorporating Variational Autoencoder (VAE) \cite{serban2017hierarchical, park2018hierarchical, du2018variational}. We plan to attempt these models in future work.

We expect to apply our trained model to various text-based psychotherapy applications, such as extracting and summarizing counseling dialogues or using the information to build a model addressing the privacy issue of training data. We hope our categorization scheme and our ConvMFiT model become a stepping stone for future computational psychotherapy research.

\section*{Ethical Approval}
The study reported in this paper was approved by the KAIST Institutional Review Board (\#IRB-17-95).

\section*{Acknowledgments}
This research was supported by the Engineering Research Center Program through the National Research Foundation of Korea (NRF) funded by the Korean Government MSIT (NRF-2018R1A5A1059921). 

\clearpage
\bibliography{naaclhlt2019}
\bibliographystyle{acl_natbib}

%\clearpage
%\input{Appendix}

\end{document}